\pdfoutput=1

\documentclass[11pt]{article}

\usepackage{ACL2023}

\usepackage{times}
\usepackage{latexsym}
\usepackage{amssymb}
\usepackage{amsmath}
\usepackage{makecell}

\usepackage[T1]{fontenc}

\usepackage[utf8]{inputenc}

\usepackage{microtype}

\usepackage{inconsolata}

\usepackage{subfigure}
\usepackage{graphicx}

\usepackage{booktabs}

\usepackage{hyperref}

\title{Following Route Instructions using Large Vision-Language Models: \\ A Comparison between Low-level and Panoramic Action Spaces}

\author{Vebjørn Haug Kåsene \\
  University of Oslo \\
  \texttt{vebjorhk@gmail.com}
  \\\And
  Pierre Lison \\
  Norwegian Computing Center \\
  \texttt{plison@nr.no} \\}

\begin{document}
\maketitle
\begin{abstract}
\textit{Vision-and-Language Navigation} (VLN) refers to the task of enabling autonomous robots to navigate unfamiliar environments by following natural language instructions. While recent Large Vision-Language Models (LVLMs) have shown promise in this task, most current VLM systems rely on models specifically designed and optimized for navigation, leaving the potential of off-the-shelf LVLMs underexplored. Furthermore, while older VLN approaches used low-level action spaces with egocentric views and atomic actions (such as "turn left" or "move forward"), newer models tend to favor panoramic action spaces with discrete navigable viewpoints. This paper investigates (1) whether off-the-shelf LVLMs (fine-tuned without architectural modifications or simulator-based training) can effectively support VLN tasks and (2) whether such models can support both low-level and panoramic action paradigms. To this end, we fine-tune the open-source model Qwen2.5-VL-3B-Instruct on the \textit{Room-to-Room (R2R)} dataset and evaluate its empirical performance across both low-level and panoramic action spaces. The best resulting model achieves a 41\% success rate on the R2R test set, demonstrating that while off-the-shelf LVLMs can learn to perform Vision-and-Language Navigation, they still lag behind models specifically designed for this task.
\end{abstract}

\section{Introduction}

\begin{figure*}[h!]
  \centering
  \begin{minipage}[t]{0.42\textwidth}
    \centering
    \includegraphics[width=\linewidth]{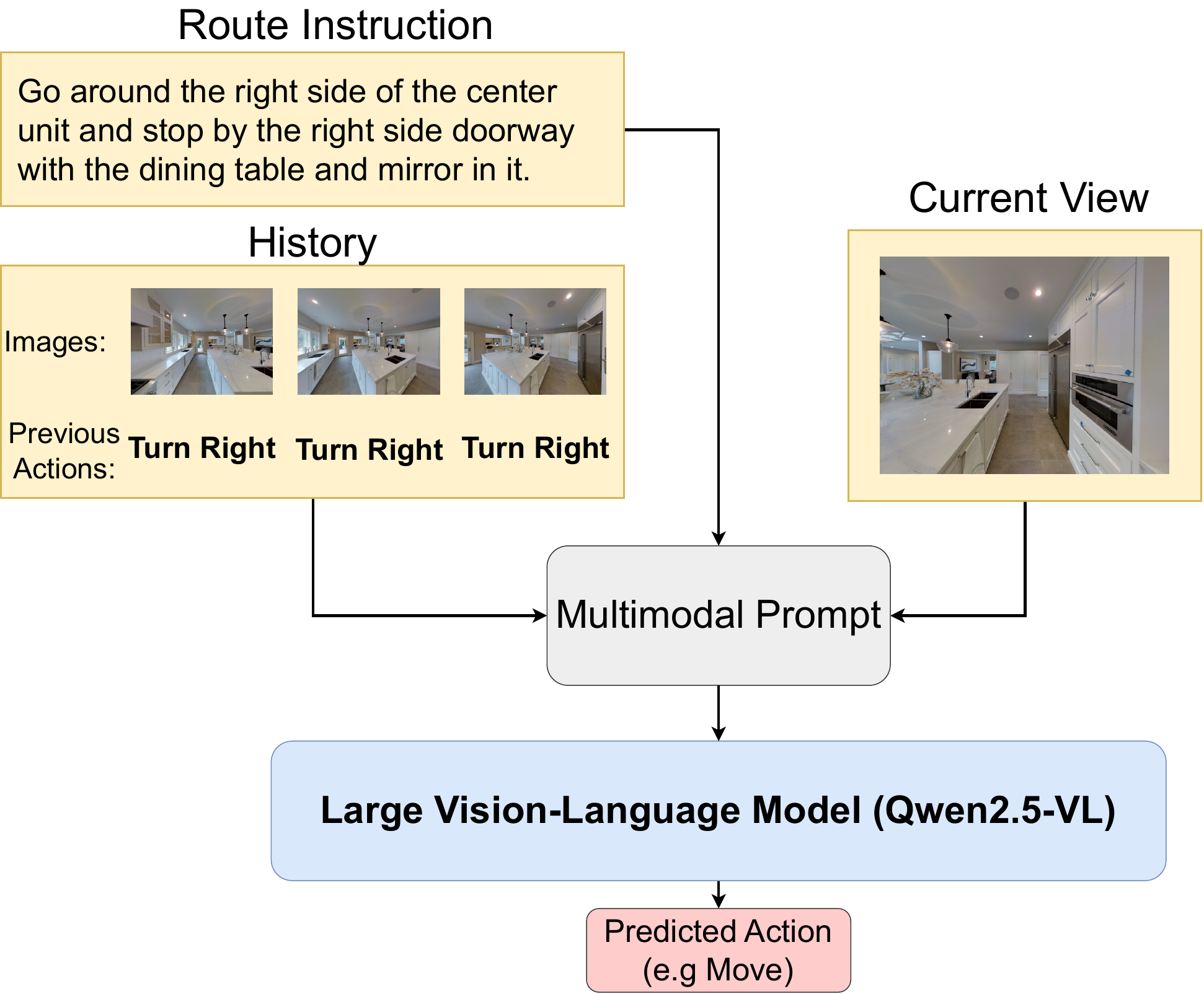}
    \caption*{(a) Low-level action space}
  \end{minipage}
  \hfill
  \begin{minipage}[t]{0.54\textwidth}
    \centering
    \includegraphics[width=\linewidth]{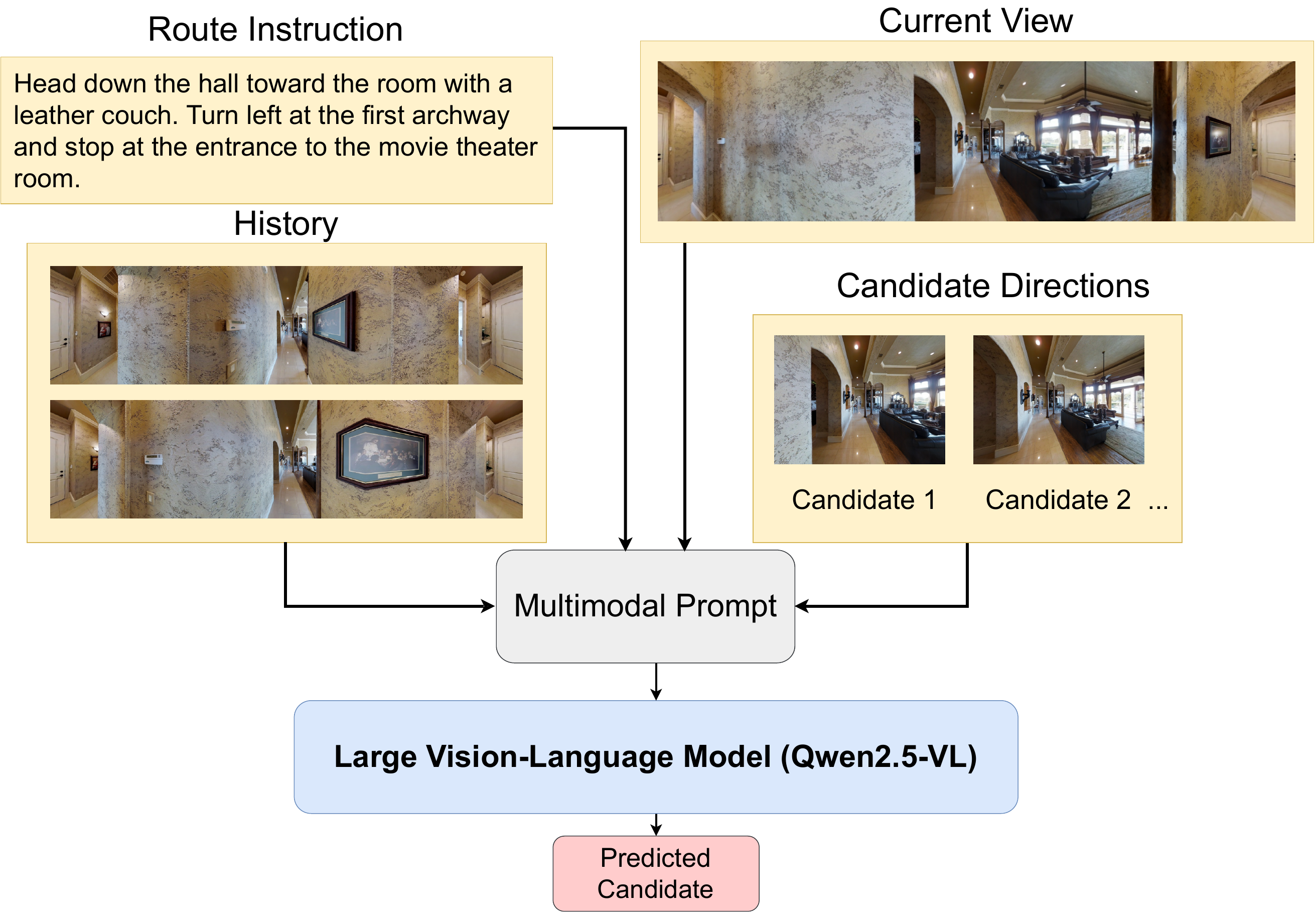}
    \caption*{(b) Panoramic action space}
  \end{minipage}
  \caption{Sketch of the approach, based on fine-tuning a pre-trained LVLM (Qwen 2.5-VL) on the R2R dataset. The LVLM receives as input a multimodal prompt including the route instruction, the navigation history and the current view, and outputs the next navigation action to perform.}
  \label{fig:overview}
\end{figure*}

Mobile robots deployed in real-world environments are often tasked with reaching specific locations described in natural language. For example, a robot might be instructed to “deliver a package to the office at the end of the hallway,” without prior knowledge of the environment. In such cases, a human can provide guidance through route instructions such as “Walk down the hallway and take the last door to your left.” To perform its task, the robot must first interpret the linguistic input provided by the human user, ground this input in its visual perception of the environment, and execute the corresponding sequence of physical actions to reach the target location.

This problem is addressed in the field of Vision-and-Language Navigation (VLN)~\citep{VLN}, which focuses on developing autonomous robotic agents that can navigate unseen environments based on natural language instructions. A common VLN benchmark and dataset is Room-to-Room (R2R)~\citep{VLN}, which contains thousands of trajectory–instruction pairs, where the task is to follow natural language instructions to reach a target location. R2R is typically used in combination with the Matterport3D simulator~\citep{VLN}, which simulates indoor environments reconstructed from real-world 3D scans from the Matterport3D dataset~\citep{matterport3d}. The simulator represents these environments as navigation graphs, where nodes correspond to navigable locations and edges define transitions between them.

Early approaches to VLN primarily relied on RNN-based sequence-to-sequence models to encode route instructions and predict actions~\citep{VLN, Speaker-Follower}. Later work shifted toward using pre-trained transformer-based models~\citep{transformers}, which offered improved language understanding and generalization~\citep{press, hamt, DUET}. 

More recently, researchers have begun exploring the use of Large Language Models (LLMs) and Large Vision-Language Models (LVLMs) for VLN, using both zero-shot prompting \citep{navGPT, MapGPT} and trained approaches \citep{NaviLLM, NavGPT-2}. While zero-shot methods have shown promise in navigation tasks, their performance still falls short of VLN-specialized transformer-based models \citep{NavGPT-2}. Most existing VLN approaches thus seek to train LLMs and LVLMs directly on VLN datasets. Although these trained approaches have achieved strong results, they typically rely on custom models that require either changes to the underlying neural architecture or the addition of task-specific components such as simulators employed at training time \citep{NaviLLM, NavGPT-2}. As a result, the potential of off-the-shelf LVLMs, fine-tuned for VLN without architectural changes, remains largely underexplored.

In addition, the choice of action space -- i.e.~the possible outputs that the model is designed to generate -- has been shown to significantly affect performance~\citep{Speaker-Follower}. Early RNN-based approaches typically employed a low-level action space, where the agent observes the environment through an egocentric image and selects from a discrete set of atomic actions such as \texttt{Move Forward}, \texttt{Turn Left}, or \texttt{Turn Right} \citep{VLN, high_level}. However, low-level action spaces have largely been abandoned in recent work in favor of panoramic action spaces~\citep{press, DUET, navGPT}, where the agent perceives its surroundings through a $360^\circ$ panoramic image and chooses among a set of navigable candidate directions, each typically corresponding to an adjacent node in the navigation graph. This shift has been shown to substantially improve performance over low-level alternatives~\citep{Speaker-Follower}. While this difference in performance has been explored in the context of RNN-based models~\citep{Speaker-Follower, high_level}, it has to our knowledge never been investigated for LVLM-based approaches.  While panoramic action spaces do seem improve the navigation performance, they also assume a greater prior knowledge about the environment -- such as which directions are navigable -- and effectively reduces the task to a visually guided graph search~\citep{high_level, VLN-CE}. Panoramic action spaces also depend on the availability of panoramic visual input, which in practice requires specialized robot-mounted hardware, such as panoramic or multi-camera rigs.

This paper seeks to address these knowledge gaps through experiments with a state-of-the-art LVLM, Qwen2.5-VL \citep{qwen2.5-VL}. An overview of our approach is illustrated in Figure~\ref{fig:overview}. The two main contributions of this paper are:
\begin{itemize}
    \item The evaluation of off-the-shelf LVLMs (without architectural changes or simulation-based training methods) on VLN through experiments on the R2R dataset. 
    
    \item An analysis of how the choice of action space (low-level versus panoramic) affects the navigation performance. 
\end{itemize}  

The rest of this paper is as follows. We first review related work on Visual-and Language Navigation and LVLMs. We then present our approach in Section 3, focusing on the fine-tuning process and the definition of possible action spaces. Section 4 then describes the experimental setup and the results obtained on the R2R dataset. Finally, Section 5 discusses those results and Section 6 concludes this paper. 

\section{Related Work}
\subsection*{Large Vision-Language Models in VLN}
Motivated by recent progress with LLMs and LVLMs, several studies have investigated how those models can be applied for VLN. NavGPT~\citep{navGPT} employs GPT-4~\citep{GPT-4} in a zero-shot setting, relying on a separate model to convert visual inputs into textual descriptions. In contrast, MapGPT~\citep{MapGPT} prompts GPT-4V to perform joint reasoning over visual inputs and navigation instructions.

NaviLLM~\citep{NaviLLM} use a frozen Vision Transformer (ViT)~\citep{ViT} and models spatial relationships between different viewpoints through a trained transformer-based multi-view fusion component which produces a single visual feature for each image. NavGPT-2~\citep{NavGPT-2} use a frozen LVLM to produce reasoning text from image-instruction pairs and fine-tunes a separate graph-based policy to predict actions and model the topological graph on the fly. Both approaches achieve state-of-the-art performance on R2R, demonstrating the potential of LLMs and LVLMs for navigation.


\subsubsection*{Action Spaces in VLN}
Early approaches to VLN employ a low-level action space where the agent perceives the world through an egocentric image at each step and predicts action such as \texttt{Move Forward} or \texttt{Turn Right} \citep{VLN, look-before-leap, Speaker-Follower}. 
\citealp{Speaker-Follower} introduce panoramic action space for VLN. Instead of receiving an egocentric image as input, the model is instead provided with a panorama comprised of 36 images a different angles. The images closest to the center of an adjacent node are considered as candidate views. Instead of predicting low-level actions, the agent instead selected between which of these views to navigate to. 
Using a LSTM~\citep{lstm} seq-2-seq model, they observe a 12\% performance increase on R2R when going from low-level to panoramic action space. 

Although there is little recent work on low-level action spaces in discrete environments (VLN-DE), it remains the most common approach for VLN in continuous environments (VLN-CE)~\citep{VLN-CE, NaVid} where agents are tasked to navigate environments not constrained by a predefined navigation graph. In this work, we focus on VLN in discrete environments. 

\subsection*{Modality alignment in LVLMs}
Modern Large Vision-Language Models (LVLMs) typically comprise three core components: a vision encoder (e.g., a Vision Transformer~\citep{ViT}), a cross-modal projector, and a text encoder (e.g, a LLM) \citep{qwen2.5-VL}. The role of the cross-modal projector is to align the visual features produced by the vision encoder with the latent space of the LLM. 

\citet{what-matters} investigate key design choices in building LVLMs and identify two prevalent architectural paradigms for vision-language alignment. The first is the \textit{cross-attention architectures}, in which visual features are injected at different layers within the LLM, an example of one such model is Flamingo~\citep{Flamingo}. The second is the \textit{fully autoregressive architectures} where the output of the vision encoder is projected into the input space of the LLM and concatenated with the sequence of text embeddings as a multimodal prompt \citep{miniGPT-4, BLIP-2}. The model used in this study, Qwen2.5-VL, follows this fully autoregressive design.

\section{Method}

\subsection{Problem Formulation}
We adopt the standard VLN in discrete environments (VLN-DE) setup \citep{VLN, Speaker-Follower, DUET}, where the environment is modeled as an undirected graph $G = \{V, E\}$. The nodes $V = \{v_i\}_{i=1}^K$ represents $K$ navigable locations while the edges $E$ constitute navigation paths between them. We then formulate the problem of following route instructions in a graph-based environment as follows: given a natural language route instruction $W = \{w_1, w_2, \dots, w_L\}$, the agent is tasked with following the instruction to reach the goal location. At each time step $t$, the agent receives a visual observation $O_t$, maintains a history context $H_t$, and is provided with auxiliary signals such as the cumulative distance traveled $ d_t \in \mathbb{R}$ and the current step number $t$. The specific formulation of the agent's input and output depends on the underlying action space, as described below.

\subsection{Low-level Action Space}
In the low-level action space, the agent perceives its environment through an egocentric image $O_t$ at each step. It maintains a historical context $H_t = \{(O_1, a_1), (O_2, a_2), \dots, (O_{t-1}, a_{t-1})\}$ where $O_{t-1}$ and $a_{t-1}$ are the image and action from the previous step, respectively. Additionally, the agent is provided with a set of low-level actions $ U_t = \{u_1, u_2, \dots, u_k \}$ that represents the actions allowed at step $t$, given the physical constraints of the environment (e.g., the agent cannot move forward if directly facing a wall). The agent predicts the next action $a_t$ by estimating the probability:
\begin{equation}
    P(a_t \mid W, O_t, H_t, d_t, t, U_t)
\end{equation}


The low-level action space used in this work consists of four discrete actions:
\begin{itemize}
    \item \texttt{Move}: moves forward to the node closest to the center of the current field of view.
    \item \texttt{Left}, \texttt{Right}: rotate the agent by $30^\circ$ in the respective direction.
    \item \texttt{Stop}: signals that the agent believes it has reached the goal.
\end{itemize}

One limitation of this setup is that navigation is constrained to a discrete graph of nodes. The \texttt{Move} action advances the agent to the node most centered in its current field of view, but this target is not necessarily aligned with the agent’s heading. As a result, the agent may appear to move sideways, which can lead to non-intuitive trajectories. To mitigate this, an automatic reorientation step, referred to as \texttt{Automatically Turn Towards Node}, is applied before each \texttt{Move} action. Although this reorientation is not part of the learnable action space, both the resulting observation and action are included in the agent’s history. This adjustment allows us to evaluate whether explicitly aligning the agent’s heading with its movement direction improves navigation performance.

\subsection{Panoramic Action Space}
With the panoramic action space, the agent perceives the environment through a $360^\circ$ panoramic image $O_t$ at each step, aligned with its current center. The agent maintains a history of panoramic views $H_t = \{O_1, \dots , O_{t-1} \}$ and selects from a set of navigable candidate views $C_t = \{c_1 \dots , c_k \}$. Each candidate $c_i$ includes an image, a relative heading $\theta_i \in [-180^\circ, 180^\circ]$, and an associated travel distance $\delta_i \in \mathbb{R}_{\geq 0}$. The task for each step is to predict the correct candidate direction $c_t$:
\begin{equation}
    P(c_t \mid W, O_t, H_t, d_t, t, C_t)
\end{equation}

Similarly to low-level actions, the episode concludes when the agent predicts the \texttt{Stop} action. 

The panoramic image is centered on the agent’s current heading, while each candidate view is a standard perspective image oriented directly toward a navigable direction. Candidate views are sorted from left to right based on their relative angle to the panoramic center, with the leftmost candidate assigned index 0 and the rightmost index $K-1$.

At each step, the model predicts a token corresponding to one of the candidate indices (from $0$ to $K{-}1$) or  the \texttt{Stop} action. Unlike traditional panoramic setups \citep{Speaker-Follower, NaviLLM}, where candidate views are extracted from within the panorama itself, this approach treats the panorama and candidate views as separate inputs. This design, motivated by memory limitations, requires fewer input images per step. See Appendix~\ref{sec:appendix} for an illustrative example. 

\subsection{Action selection}

\begin{figure*}[t]
  \centering
  \begin{minipage}[t]{0.48\textwidth}
    \centering
    \includegraphics[width=\linewidth]{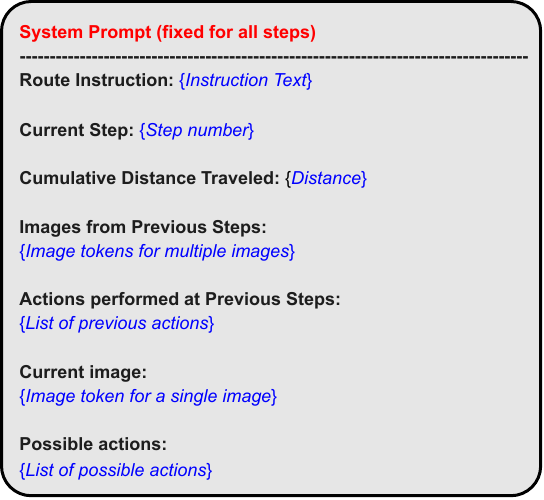}
    \caption*{(a) Low-level action space prompt schema}
  \end{minipage}
  \hfill
  \begin{minipage}[t]{0.48\textwidth}
    \centering
    \includegraphics[width=\linewidth]{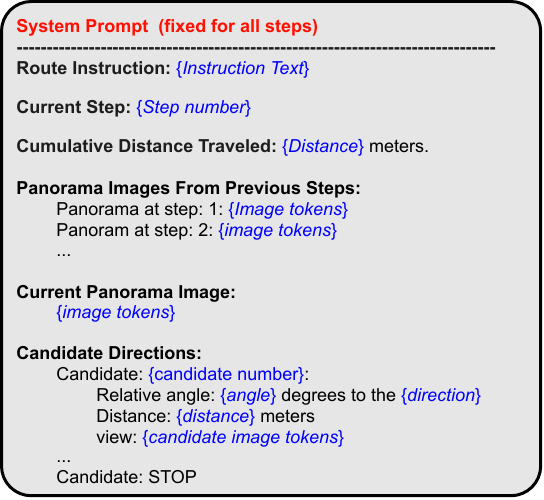}
    \caption*{(b) Panoramic action space prompt schema}
  \end{minipage}
  \caption{Prompt schemata for low-level and panoramic action spaces.}
  \label{fig:prompt-schemas}
\end{figure*}

To select the next action to perform, the model receives a structured multi-modal prompt that encodes the current state, including the instruction, visual input, and auxiliary information such as step number and distance traveled. These prompts follow a fixed schema shown in Figure~\ref{fig:prompt-schemas}.  Inference is performed greedily, selecting the most probable action at each step without backtracking.

In addition to the dynamic input state, each prompt includes a static system prompt that explains the task and describes the individual input fields. The system prompt is fixed and specific to each action space, and remains unchanged throughout training and evaluation. The full system prompts are included in Appendix~\ref{sec:appendix}

\subsection{Fine-tuning}

The LVLMs are fine-tuned through behavior cloning, where the model learns to imitate expert demonstrations. At each time step $t$, the model receives a multimodal prompt $x_t$ represent the current state, and is trained to predict the expert action $a_t$ as a token from its own vocabulary. The training objective minimizes the total negative log-likelihood of the expert actions over the entire episode.
Gradients are accumulated across all time steps in an episode, and the weights are updated at the end of each episode. 

Unlike many recent VLN approaches~\citep{hamt, NavGPT-2, VLN}, our approach does not therefore rely on reinforcement learning or student forcing, but simply fine-tunes the LVLM model on the basis of expert routes. A key advantage of this approach is the fact that it can be applied without access to a simulator at training time.  

\section{Experiments}

The proposed approach was evaluated on the Room-to-Room (R2R) dataset using both \textit{offline} and \textit{online} evaluation modes. The offline mode assesses the model’s ability to follow expert trajectories, whereas the online mode evaluates its performance when navigating autonomously within the Matterport3D simulation environment.

\subsection{Dataset}

The Room-to-Room dataset~\citep{VLN} contains 21,567 English route instructions corresponding to 7,189 trajectories across 90 environments. Each ground truth trajectory is a sequence of nodes in a Matterport3D environment. Each trajectory has 3 corresponding instructions. 

The dataset is split into four subsets: \textit{train} (61 environments), \textit{val seen} (56 environments overlapping with train), \textit{val unseen} (11 environments), \textit{and test} (18  environments). Performance is evaluated on the val unseen and test splits. All splits are preprocessed to convert ground truth trajectories into sequences of actions. \footnote{For the alternative low-level action spaces experiments, the models were trained on a subset consisting of the first 1,955 trajectories in R2R's train split.}





\subsection{Evaluation Metrics}
Online, the models are evaluated using standard VLN metrics \citep{VLN}. \textbf{Navigation Error (NE)} is the average walkable distance between the agent's final location and the goal location in meter. An episode is considered successful if NE $\leq 3$ m and the last predicted action is \texttt{Stop}. \textbf{Path Length (PL)} is the average path length (in meters).  \textbf{Oracle Success Rate (OSR)} measures the percentage of episodes in which the agent was within 3 meters of the goal at any point during the navigation episode. \textbf{Success Rate (SR)} is the percentage of episodes that are successful. \textbf{Success Weighted by Path length (SPL)}~\citep{eval_important} combines SR with path length, penalizing unnecessarily long paths. \textbf{Coverage Weighted by Length Score (CLS)}~\citep{r4r} measures how well the agent's predicted path follows the route instruction, penalizing paths which deviate from the ground truth path.

For the offline evaluation, the reported metrics are \textbf{Accuracy} the proportion of actions correctly predicted by the model; \textbf{Macro F1}, the unweighted mean F1 score computed across all action classes; and \textbf{Conservative Success Rate (CSR)}, the percentage of episodes in which all actions are identical (from start to finish) to the actions selected by the expert. For offline evaluation, we use the third instruction from each trajectory.

\subsection{Implementation}

\subsubsection*{Model}
We experimented with two distinct Large Vision-Language Models (LVLMs): Qwen2-VL-2B-Instruct~\citep{qwen2-VL} and Qwen2.5-VL-3B-Instruct~\citep{qwen2.5-VL}. Qwen2.5-VL-3B is the larger of the two and is pre-trained on 4 trillion tokens, compared to 1.2 trillion tokens for Qwen2-VL-2B. As our experiments showed that Qwen2.5 consistently provided superior performance compared to Qwen2 in both offline and online metrics on the validation dataset, we only provide evaluation results obtained for Qwen2.5.  

During fine-tuning, the vision encoder and the cross-modal projection layer are kept frozen, as preliminary experiments indicated that tuning only the LLM led to improved performance.\footnote{The trained models are publicly available at \url{https://huggingface.co/Vebbern} for reproducibility.} 

\subsubsection*{Simulator}
The Matterport3D simulator (MP3D) is used for evaluation and for generating the preprocessed training data. In MP3D, the agent’s field of view is determined by the image resolution and the vertical field of view (VFOV). This work uses an image resolution of 640$\times$480 for egocentric and candidate images. The VFOV is set to $105^\circ$ to allow the agent to perceive a broader visual context. This is substantially larger than the VFOV used in prior work, which typically ranges from $60^\circ$~\citep{VLN, Speaker-Follower} to $75^\circ$~\citep{VLN-CE}. Panoramic images are constructed by stitching together three egocentric views captured while rotating the agent in place.

\subsubsection*{Training} 
All models are fine-tuned with a batch size of 1, a learning rate of 1e-5, and a weight decay of 0.1. A linear learning rate scheduler is used with warmup over the first 10\% of training steps. FlashAttention~\citep{FlashAttention} is enabled, and training is performed in bfloat16 precision. Input images are resized to half their original size to accommodate GPU memory constraints.\footnote{Meaning 320$\times$240 for candidates and egocentric views, and 960$\times$240 for panoramic views} Experiments were conducted on a single NVIDIA A100 80GB GPU. Models are fine-tuned for 1 epoch across all instructions, corresponding to 3 total passes over the unique paths (as each path in R2R is associated with three distinct route instructions). \footnote{The source code is available at \url{https://github.com/Vebjorhk/masters-thesis-VLN}.} 

\subsection{Results}

\paragraph{Offline evaluation}
Table ~\ref{tab:r2r-full-offline} presents the offline evaluation results after fine-tuning the Qwen2.5 model on the full training set of R2R.  In terms of accuracy, the panoramic and low-level models scores similarly. The low-level model has a slightly lower macro F1 score, which could be explained by the larger number of actions of panoramic models (up to 12 candidate views). However, the panoramic model have a significantly higher conservative success rate (SCR) than the low-level one. Qwen2.5-VL-pano achieves a CSR of 15\% on val unseen, compared to a mere 3\% CSR for Qwen2.5-VL-low.

\begin{table}
    \centering
    \resizebox{\columnwidth}{!}{
        \begin{tabular}{lccc}
            \toprule
            \textbf{Model}
            & \textbf{Accuracy↑}
            & \textbf{Macro F1↑}
            & \textbf{CSR↑} \\
            \midrule
            \textbf{Val seen:} & & &  \\
            Qwen2.5-VL-low & 0.73 & 0.74 & 0.04 \\
            Qwen2.5-VL-pano & 0.73 & 0.61 & 0.16 \\
            \midrule
            \textbf{Val unseen:} & & &  \\
            Qwen2.5-VL-low & 0.73 & 0.73 & 0.03 \\ 
            Qwen2.5-VL-pano & 0.73 & 0.62 & 0.15 \\
            \bottomrule
        \end{tabular}
    }
    \caption{Offline evaluation results on the seen and unseen R2R validation sets.}
    \label{tab:r2r-full-offline}
\end{table}

\begin{table*}
    \small
    \renewcommand{\arraystretch}{1.3}
    \setlength{\tabcolsep}{5pt}
    \resizebox{\textwidth}{!}{
        \begin{tabular}{lcccccccccccccccc}
        \toprule
        & \multicolumn{5}{c}{\textbf{Val Seen}} 
        & \multicolumn{5}{c}{\textbf{Val Unseen}} 
        & \multicolumn{5}{c}{\textbf{Test (Unseen)}} \\
        & PL & NE↓ & OSR↑ & SR↑ & SPL↑ 
        & PL & NE↓ & OSR↑ & SR↑ & SPL↑ 
        & PL & NE↓ & OSR↑ & SR↑ & SPL↑ \\
        \midrule
        Human
        & - & - & - & - & - 
        & - & - & - & - & -
        & 11.85 & 1.61 & 90 &  86 & 76 \\
        \midrule
        \textbf{Low-Level}
        &  &  &  &  & 
        &  &  &  &  & 
        &  &  &  &  &  \\
        
        Seq2Seq~\shortcite{VLN}
        & 11.33 & 6.01 & 53 & 39 & - 
        & 8.39 & 7.81 & 28 & 22 & -
        & 8.13 & 7.85 & 27 & 21 & - \\

        SF~\shortcite{Speaker-Follower} 
        & - & 4.28 & 60 & 47 & - 
        & - & 5.75 & 33 & 25 & -
        & - & - & - & - & - \\

        RPA\shortcite{look-before-leap} 
        & 8.46 & 5.56 & 53 & 43 & - 
        & 7.22 & 7.65 & 32 & 25 & -
        & 9.15 & 7.53 & 33 & 25 & - \\

        DCF~\shortcite{high_level} 
        & - & 3.96 & 73 & 58 & 51 
        & - & 6.52 & 43 & 34 & 29
        & 9.81 & 6.55 & 45 & 35 & 31 \\
        
        \midrule
        \textbf{Panoramic}
        &  &  &  &  & 
        &  &  &  &  & 
        &  &  &  &  &  \\

        SF~\shortcite{Speaker-Follower} 
        & - & 3.36 & 73 & 66 & - 
        & - & 6.62 & 45& 36 & -
        & - & - & - & - & - \\

        PRESS~\shortcite{press} 
        & 10.35 & 3.09 & - & 71 & 67 
        & 10.06 & 4.31 & - & 59 & 55
        & 10.52 & 4.53 & 63 & 57 & 53 \\

        VLN~\ensuremath{\circlearrowright}~BERT~\shortcite{VLN_BERT}
        & 11.13 & 2.90 & - & 72 & 68 
        & 12.01 & 3.93 & - & 63 & 57
        & 12.35 & 4.09 & - & 63 & 57 \\

        HAMT~\shortcite{hamt}
        & 11.15 & 2.51 & - & 76 & 72 
        & 11.46 & 2.29 & - & 66 & 61
        & 12.27 & 3.93 & - & 65 & 60 \\

        DUET~\shortcite{DUET} 
        & - & - & - & - & - 
        & 13.94 & 3.31 & - & 72 & 60
        & 14.73 & 3.65 & - & 69 & 59 \\

        NavGPT~\shortcite{navGPT}
        & - & - & - & - & - 
        & - & - & - & - & -
        & 11.45 & 6.46 & 42 & 34 & 29 \\

        NaviLLM~\shortcite{NaviLLM}
         & - & - & - & - & - 
        & - & - & - & - & 59
        & - & - & - & - & 60 \\

        NavGPT-2~\shortcite{NavGPT-2}
        & 14.13 & 2.84 & 83 & 74 & 63
        & 14.01 & 2.98 & 84 & 74 & 61 
        & 14.74 & 3.33 & 80 & 72 & 60 \\
        
        \midrule
        \textbf{Qwen2.5-VL-low} 
        & 10.27 & 7.14 & 41 & 35 & 32 
        & 10.50 & 7.84 & 34 & 27 & 24 
        & 10.59 & 7.99 & 34 & 26 & 24 \\
        
        \textbf{Qwen2.5-VL-pano} 
        & 9.98 & 5.69 & 56 & 50 & 47 
        & 9.83 & 6.65 & 46 & 38 & 35 
        & 9.96 & 6.53 & 50 & 41 & 38 \\
        \bottomrule
    \end{tabular}
    }
    \caption{Comparison of panoramic and low-level models with state-of-the-art performance using single-run greedy search. R2R does not report CLS. The models presented in this work is in bold text.}
    \label{tab:r2r-sota}
\end{table*}

\paragraph{Online evaluation}
Table~\ref{tab:r2r-sota} compares our results with state-of-the-art (SOTA) approaches on R2R using single-run greedy search (i.e., no pre-exploration). Results are shown for both panoramic and low-level action space.

The model fine-tuned for low-level action spaces, Qwen2.5-VL-low, achieves a success rate (SR) of 26\% on the test set, outperforming the original R2R baseline (Seq2Seq, 21\% SR) and Speaker-Follower (SF) without panoramic action (25\% SR on val unseen). However, it still lags behind the LSTM-based DCF model of \cite{high_level}, which reached 35\% SR, despite being substantially smaller in size. However, Qwen2.5-VL-low is less prone to overfitting to training environments, as reflected in the smaller SR gap between val seen and unseen (35\% vs. 27\%) compared to DCF, which drops from 58\% to 34\% on val unseen.

The panoramic model, Qwen2.5-VL-pano, achieves a 41\% SR on the R2R test set. This outperforms all low-level models as well as panoramic approaches such as Speaker-Follower (36\% on val unseen) and NavGPT~\citep{navGPT} (34\% on test). However, this model falls short of more recent task-specific panoramic approaches such as NaviLLM~\citep{NaviLLM} (60\% SPL) and NavGPT-2~\citep{NavGPT-2} (72\% SR).

\paragraph{Alternative low-level action spaces}
We also explored alternative configurations for the low-level action space. Specifically, we assessed the  impact of (1) disabling the \texttt{Automatically Turn Towards Node} action, and (2) reducing the vertical field of view (VFOV) from $105^\circ$ to a narrower $82^\circ$. 

Table~\ref{tab:alt-data-MP3D} presents the performance on the val unseen split for those two alternatives. The difference between $82^\circ$ and $105^\circ$ VFOV is minimal, with only slight improvements in CLS and OSR scores for the $105^\circ$ configuration. However, removing the adjustment action leads to a noticeable performance gain: the No-Adjust model achieves a SR of 29\%, compared to 25\% for the default. This suggests that explicitly facing the next node before movement is often unnecessary for effective navigation. 


\begin{table}[t]
    \centering
    \resizebox{\columnwidth}{!}{%
    \begin{tabular}{lcccccc}
        \toprule
        \textbf{Models} 
        & \textbf{PL} 
        & \textbf{NE↓}  
        & \textbf{OSR↑}
        & \textbf{SR↑}
        & \textbf{SPL↑}
        & \textbf{CSL↑} \\
        \midrule
        \textbf{Val Unseen:} & & & & & &  \\
        105-VFOV & 10.17 & 7.87 & 0.34 & 0.25 & 0.23 & 0.45\\
        82-VFOV &  9.9 & 7.87 & 0.32 & 0.25 & 0.23 & 0.44 \\
        \textbf{No-Adjust} & 10.72 & 7.7 & 0.38 & 0.29 & 0.26 & 0.44 \\
        \bottomrule
    \end{tabular}
    }
    \caption{Online results on R2R val unseen for alternative definitions of the low-level action space. } 
    \label{tab:alt-data-MP3D}
\end{table}
\section{Discussion}

\paragraph{Fine-tuning off-the-shelf LVLMs for R2R} The results indicate that fine-tuning off-the-shelf LVLMs such as Qwen2.5-VL on the R2R task fails to yield strong performance, despite the fact that such models are significantly larger than older, VLN-specific architectures such as PRESS~\citep{press}, DUET~\citep{DUET}, and HAMT~\citep{hamt}. It is difficult to pinpoint the exact source of this performance gap, though our use of behavior cloning -- rather than optimisation through student forcing and/or reinforcement learning, as done by e.g.~\citep{hamt, NavGPT-2} -- may be a contributing factor. 

Compared to NaviLLM~\citep{NaviLLM} and NavGPT-2~\citep{NavGPT-2}, which are the two approaches most similar to this work, a key difference becomes apparent. In the Qwen2.5-VL-low model, each input image is encoded and then fed directly into the LLM, which is solely responsible for interpreting the route instruction, modeling spatial relationships between images, and predicting actions. While Qwen2.5-VL reduces visual token count through patch merging, it does not incorporate any explicit mechanisms for modeling spatial structure between images before they are fed to the LLM. In contrast, NaviLLM~\citep{NaviLLM} includes a transformer-based module that explicitly captures the spatial relationships between panoramic images before it is fed to as input to the LLM. NavGPT-2~\citep{NavGPT-2} takes this further by using a separate graph-based policy network to model viewpoint connectivity and predict actions, while delegating route-level reasoning to the LLM. These design differences may help explain at least part of the observed performance gap: relying solely on the LLM for spatial reasoning and control can be challenging -- especially for longer paths -- compared to models that explicitly encode spatial structure. Prior work also suggests that reducing visual tokens benefits non-OCR tasks~\citep{what-matters}. Both NaviLLM and NavGPT-2 use significantly fewer visual tokens than Qwen2.5-VL~\citep{qwen2.5-VL}.

\begin{table}[t]
    \centering
    \normalsize  
    \begin{tabular}{lcc}
        \toprule
        \textbf{Split} 
        & \shortstack{\textbf{Avg. steps} \\ \textbf{per path} \\ \textbf{(low-level)}} 
        & \shortstack{\textbf{Avg. steps} \\ \textbf{per path} \\ \textbf{(panoramic)}} \\
        \midrule
        train & 12.88 & 6.00 \\
        val seen & 12.85 & 6.07 \\
        val unseen & 13.40 & 5.97 \\
        \bottomrule
    \end{tabular}
    \caption{Average number of steps (actions) for panoramic and low-level variants of R2R.}
    \label{tab:r2r-stats}
\end{table}

\begin{figure}[t]
    \centering
    \includegraphics[width=1.0\linewidth]{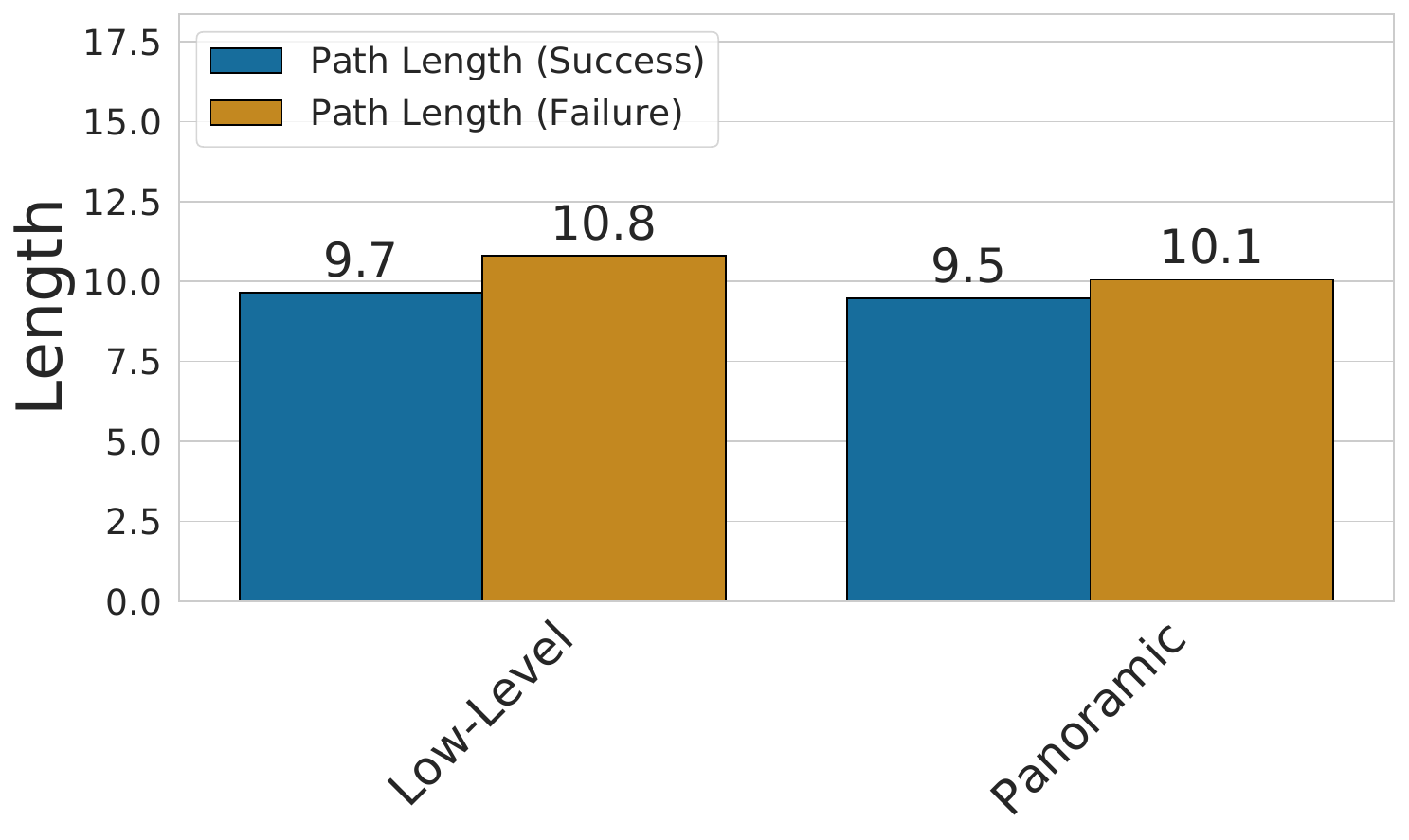}
    \caption{Avg. path length (meters) on R2R val unseen.}
    \label{fig:path-length-unseen}
\end{figure}

\paragraph{Panoramic vs. low-level action space} The panoramic models consistently outperform low-level ones, which aligns with previous findings by \citet{Speaker-Follower}, although the performance gap in our setup is slightly larger (16\% vs. 12\% SR) showing that the panoramic approach leads to better results for LVLMs as well. One plausible explanation for the performance gap is that low-level action sequences are, on average, twice as long as those in the panoramic setting (Table~\ref{tab:r2r-stats}). As shown in Figure~\ref{fig:path-length-unseen}, both model types tend to perform better on shorter trajectories. This suggests that longer sequences in the low-level setting increase the difficulty of the task, as they provide more opportunities for errors to accumulate and make recovery more challenging. This is further supported by the noticeably higher Conservative Success Rate (CSR) for panoramic models (Table~\ref{tab:r2r-full-offline}), indicating they are more likely to keep the agent on the correct path. In contrast, low-level models are more prone to errors due to the increased number of decision points, making it harder to recover once the agent deviates from the intended path.

The extent to which the additional visual information provided by panoramic images contributes to improved performance remains somewhat unclear. Panoramic observations may benefit the agent by reducing the need for physical reorientation to perceive important landmarks. Low-level action spaces may also place greater demands on spatial and temporal reasoning abilities: the agent must not only ground instructions in the visual context but also anticipate when certain actions should be executed -- such as recognizing that a given action may only occur after completing several turns.

\section{Conclusion}
This work focused on the use of off-the-shelf Large Vision-Language Models (LVLMs) for  Vision-and-Language Navigation (VLN) tasks. More precisely, we investigated how such models could be fine-tuned directly from expert routes, without modifying the model's underlying architecture or relying on online approaches that necessitate the use of a simulator at training time. The performance of this approach was assessed through experiments on the R2R dataset and explored using both panoramic and low-level action spaces. 

The best performing model, fine-tuned from Qwen2.5-VL, achieved a success rate (SR) of 41\% on the R2R dataset. Our results suggest that simply fine-tuning LVLMs remains insufficient to reach state-of-the-art performance on navigation tasks. Furthermore, we find that the performance gap between low-level and panoramic action spaces persists even with larger and more powerful models, with a 16\% difference in SR on the R2R test set in favor of the panoramic setup.


A promising topic for future work is the systematic study of off-the-shelf LVLMs on the R2R dataset. Evaluating a broader range of models beyond Qwen2 and Qwen 2.5 could help identify which architectural choices lead to better navigation performance. Additionally, a more focused investigation of the panoramic action space is warranted -- particularly through ablation studies that isolate the effect of including a panoramic view, and systematically vary the field of view to understand its impact on performance. We also encourage future work to further explore the low-level action space for more recent approaches, including adapting it to existing state-of-the-art methods such as NaviLLM~\citep{NaviLLM} and NavGPT-2~\citep{NavGPT-2} and comparing the performance to panoramic action space.

\section*{Limitations}
We acknowledge several limitations in this work. Most importantly, the fine-tuning approach is limited to behavior cloning, and did not include the use of VLN training techniques such as student forcing or reinforcement learning, potentially limiting direct comparability with prior work that leverages these strategies. For evaluation, we set up a web API to communicate with the machine running the simulator remotely. However, this introduced an additional limitation: the need for network calls made simulator evaluation significantly more time-consuming. As a result, we restricted evaluation of alternative setups to only the first third of the route instructions.

GPU memory limitations restricted training to a batch size of 1. To further reduce memory usage, we deviated from the standard panoramic action format used in many VLN approaches~\citep{Speaker-Follower, press, VLN_BERT}, where the model receives a set of discrete view images (typically 36) and selects candidate views from among them. Instead, we preprocessed the full panorama as a single image and treated candidate views as separate, independent inputs. This setup limits direct comparability, as the granularity and spatial alignment of visual information differ from the standard formulation. Additionally, the preprocessed panoramas used in this work are only roughly stitched together, which introduces some visual artifacts and further distinguishes our input format from existing benchmarks.

Finally, we note that Room-to-Room (R2R) contains only English-language route instructions, which limits the applicability of our approach to English-only scenarios. While multilingual VLN datasets have been proposed -- such as Room-across-Room (RxR)~\citep{rxr} -- our current experiments do not address multilingual aspects.


\section*{Ethics Statement}
This work investigated the use of off-the-shelf Large Vision-Language Models (LVLMs) for Vision-and-Language Navigation (VLN) tasks. All models used in this study are open-source and publicly available. The dataset employed, Room-to-Room (R2R), is a widely used benchmark in the VLN community and does not contain personally identifiable information. We do not foresee any direct ethical concerns related to the methods, data, or potential applications of this research. Our study adheres to the ACL Code of Ethics.

\section*{Acknowledgments}
This work is based on research conducted as part of the first author's Master’s thesis at the University of Oslo in 2025. Pierre Lison's work was funded by the SFI NorwAI, (Center for Research-based Innovation, 309834). All figures and visualizations derived from the Matterport3D dataset are used in accordance with the Matterport Terms of Service: \url{https://kaldir.vc.in.tum.de/matterport/MP_TOS.pdf}.

\bibliography{references}
\bibliographystyle{acl_natbib}

\appendix

\section{Appendix}
\label{sec:appendix}

\begin{figure*}[t]
    \centering
    \includegraphics[width=1.0\linewidth]{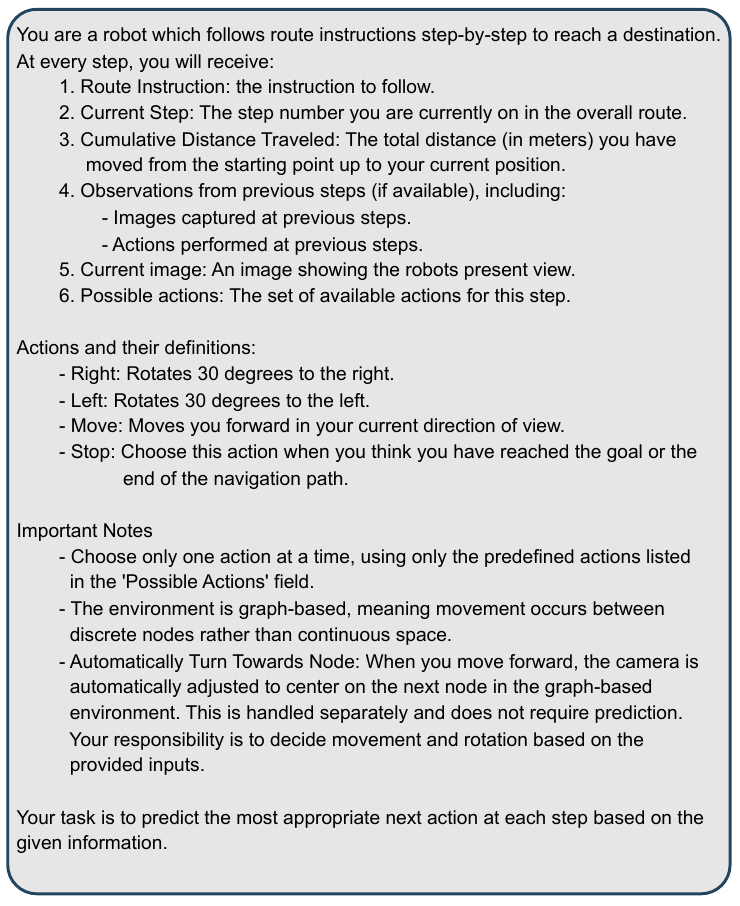}
    \caption{System prompt for low-level action space}
    \label{fig:system-prompt-low-level}
\end{figure*}

\begin{figure*}[t]
    \centering
    \includegraphics[width=1.0\linewidth]{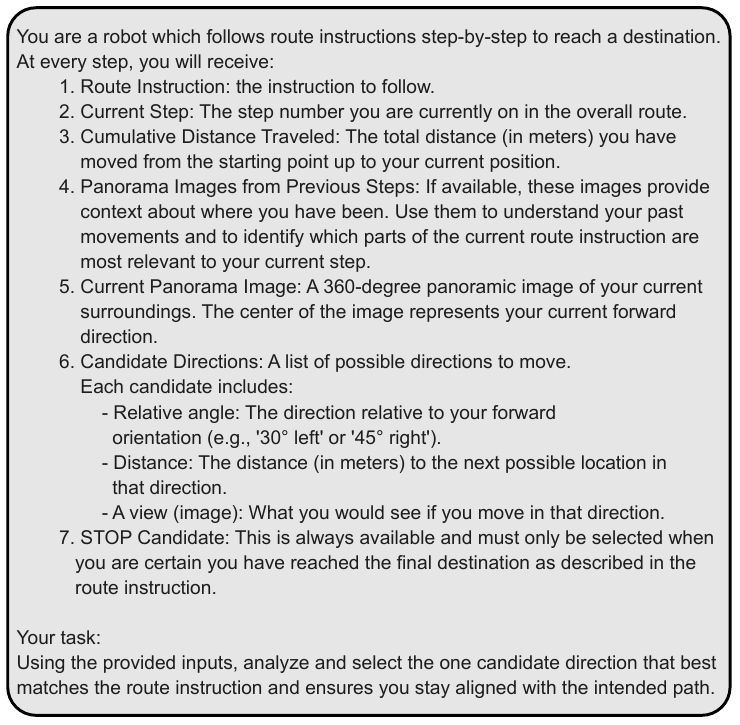}
    \caption{System prompt for panoramic action space}
    \label{fig:system-prompt-panoramic}
\end{figure*}

\begin{figure*}[t]
    \centering
    \includegraphics[width=1.0\linewidth]{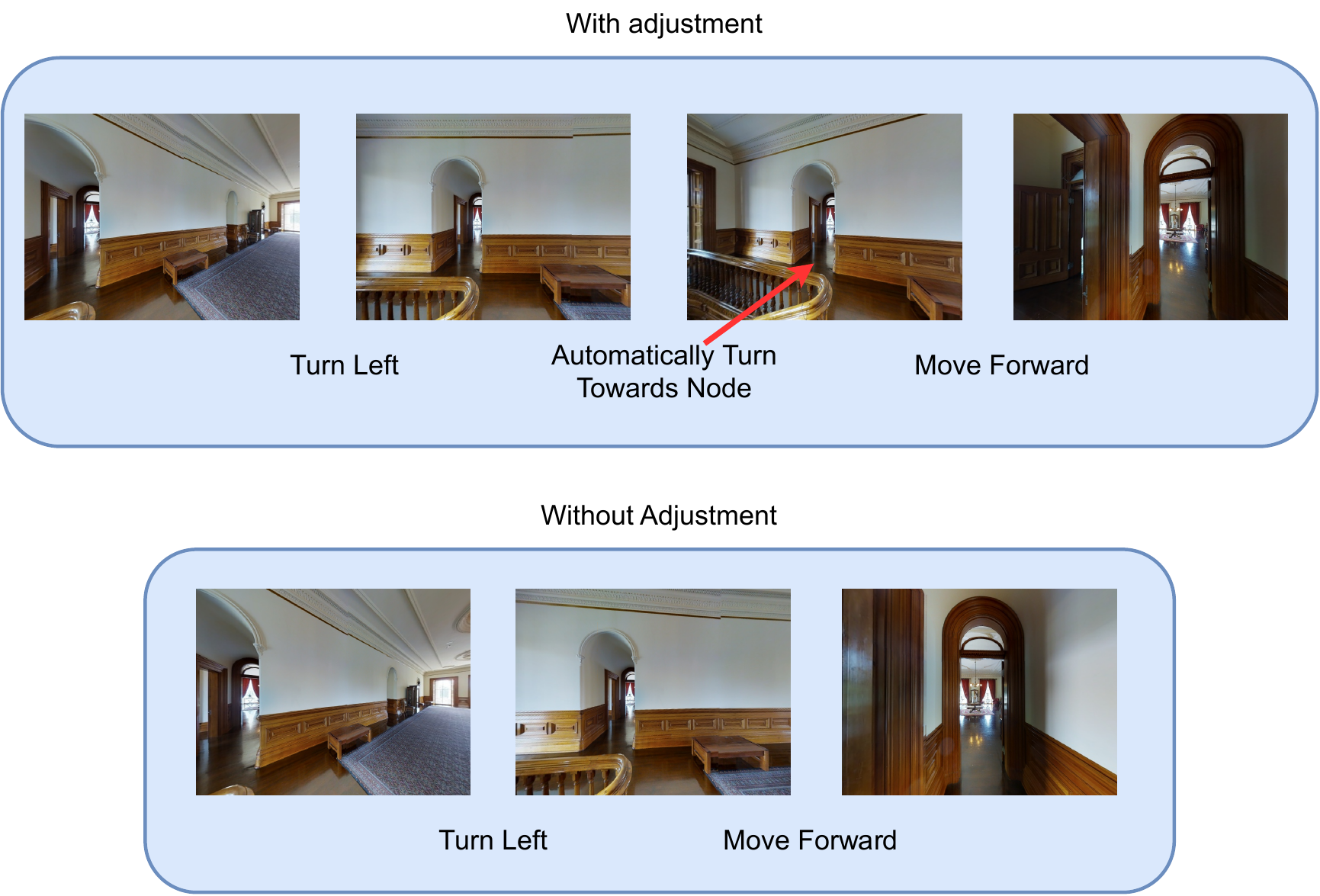}
    \caption{Figure illustrating the \texttt{Automatically Turn Towards Node} step.}
    \label{fig:adjust-example}
\end{figure*}

\begin{figure*}[t]
    \centering
    \includegraphics[width=1.0\linewidth]{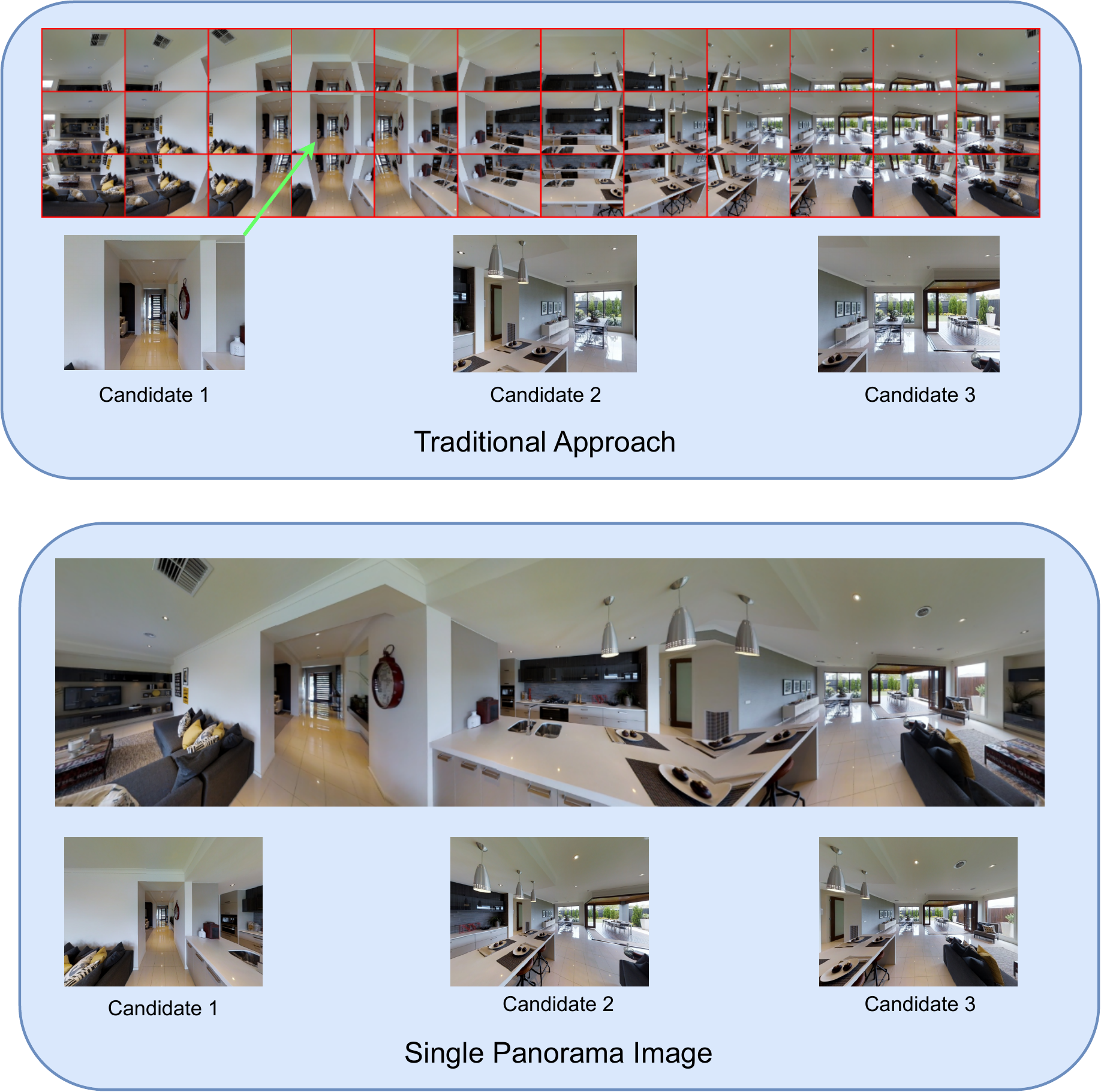}
    \caption{Figure illustrating the difference between the traditional panoramic approach and our implementation}
    \label{fig:custom-vs-traditional}
\end{figure*}
\end{document}